\documentclass[letterpaper, 10 pt, conference]{ieeeconf}  

\IEEEoverridecommandlockouts                              

\usepackage{graphics} 
\usepackage{amsmath} 
\usepackage{amssymb}  
\usepackage{acronym}
\usepackage{psfrag}  
\usepackage[process=auto]{pstool}
\usepackage{bm}
\usepackage{graphicx} 
\usepackage[font=small,labelfont=bf]{caption} 
\usepackage{subcaption} 

\usepackage{algorithmic}
\usepackage{algorithm}
\usepackage{url}

\usepackage{lipsum,graphicx,subcaption}

\title{\LARGE \bf
Nonlinear disturbance attenuation control of hydraulic robotics
}
\author{Peng Lu$^{1}$, Timothy Sandy$^{2}$ and Jonas Buchli$^{2}$
\thanks{$^{1}$Interdisciplinary Division of Aeronautical and Aviation Engineering, Hong Kong Polytechnic University, HKSAR, China
        {\tt\small peng.lu@polyu.edu.hk} }%
\thanks{$^{2}$Agile \& Dexterous Robotics Lab at the Institute of Robotics and Intelligent Systems, ETH Z{\"u}rich, Switzerland
        {\tt\small \{tsandy and buchlij\}@ethz.ch} }%
}

\begin{document}

\acrodef{SLQ}{Sequential Linear Quadratic}
\acrodef{EKF}{Extended Kalman Filter}
\acrodef{UKF}{Unscented Kalman Filter}
\acrodef{AUKF}{Adaptive Unscented Kalman Filter}
\acrodef{DOF}{Degrees-Of-Freedom}
\acrodef{RMSE}{Root-Mean Squared errors}

\acrodef{SAA}{Shoulder Adduction/Abduction}
\acrodef{SFE}{Shoulder Flexion/Extension}
\acrodef{HR}{Humerus Rotation}
\acrodef{EFE}{Elbow Flexion/Extension}
\acrodef{WR}{Wrist Rotation}
\acrodef{WFE}{Wrist Flexion/Extension}

\maketitle
\thispagestyle{empty}
\pagestyle{empty}

\begin{abstract}
This paper presents a novel nonlinear disturbance rejection control for hydraulic robots.
This method requires two third-order filters as well as inverse dynamics in order to estimate the disturbances. All the parameters for the third-order filters are pre-defined. The proposed method is nonlinear, which does not require the linearization of the rigid body dynamics. The estimated disturbances are used by the nonlinear controller in order to achieve disturbance attenuation. The performance of the proposed approach is compared with existing approaches. Finally, the tracking performance and robustness of the proposed approach is validated extensively on real hardware by performing different tasks under either internal or both internal and external disturbances. The experimental results demonstrate the robustness and superior tracking performance of the proposed approach.
\end{abstract}




\section{INTRODUCTION}
\label{s:1}
Model-based control \cite{KHALIL2002,Slotine1991} has received a lot of interesting applications due to the exploitation of the mathematical model of the system and has been used in many robotic platforms \cite{Sciavicco2005,Craig2005}. This control technique has shown excellent performance compared to traditional linear controllers such as PID control. As an example, aircraft are able to perform more aggressive maneuvers using model-based controllers, which would be difficult to achieve using linear controllers due to significant nonlinearities in the system dynamics \cite{Lu2016b}. 

However, model-based control suffers from model uncertainties. Since it is designed based on the model of the system, if the system model is inaccurate, wrong actions could be generated by the controller. To deal with that, many nonlinear adaptive control methods \cite{KHALIL2002,Slotine1986,Mistry2010,Lu2016b} are proposed. These adaptive control methods significantly improve the robustness of nonlinear control in the presence of model uncertainties. Many of these nonlinear \cite{Khatib1987,Nakanishi2008,Mistry2010} and adaptive control methods \cite{Slotine1986} are applied to robots .

\begin{figure}
\centering
\includegraphics[width = 0.32\textwidth]{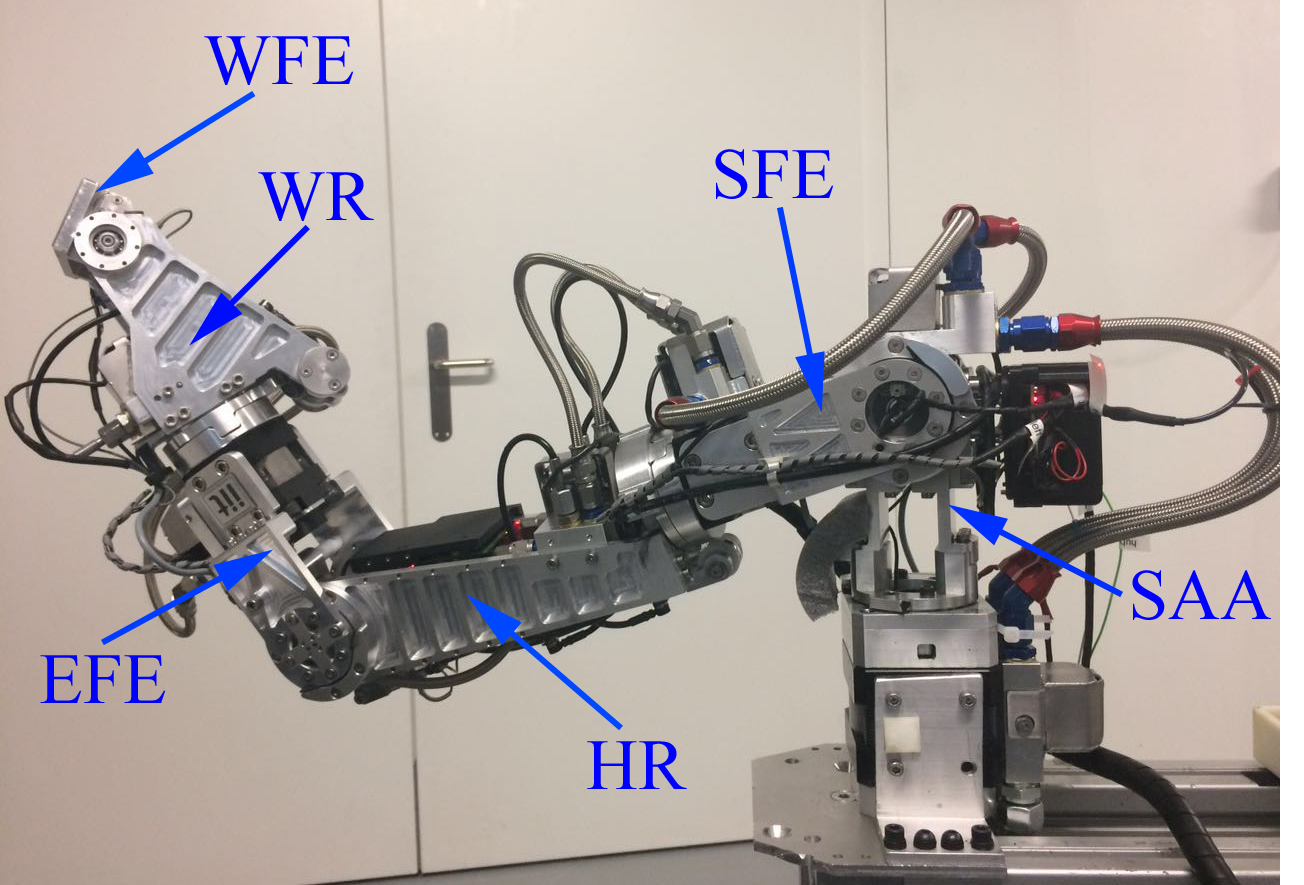}
\caption{The hydraulic arm (HyA) used in the experiments. It is actuated by hydraulic actuators which suffer from significant friction.}
\label{f:hya}
\end{figure}

Besides model uncertainties, disturbances including both internal and external disturbances pose an even greater challenge to model-based control techniques. Internal disturbances include friction forces in the actuators. For the robotic arm shown in Fig.~\ref{f:hya}, static and sliding friction forces can degrade the tracking performance of the actuators, thus decreasing the performance of the robotic arm.
The behaviour of external disturbances is unknown and difficult to predict. It is also difficult to predict when these external disturbances happen. Therefore, these external disturbances need to be taken into account to enhance the safety and performance of the robots. 

Various disturbance attenuation control methods have been proposed. These include disturbance observers \cite{Chen2000,Mohammadi2013,Kim2015,Oh2016}, sliding mode control \cite{Utkin1977,Spurgeon2014} and robust control \cite{Skogestad2001,Chen2008}. Most of them are designed based on linearized models of the rigid body dynamics, which could be computationally expensive especially when the \ac{DOF} of the robot is high. Disturbance observers usually ignore the noise in the sensors. Sliding model control does not require the linearization of the system model, but it requires the knowledge of the bound of the disturbances and its derivatives which could be difficult to obtain. In addition, sliding mode control does not consider the influence of noise. In addition, robust control is designed by considering all cases of uncertainties and disturbances, which makes it fairly conservative when there are no uncertainties and disturbances \cite{Skogestad2001}.

In this paper, we propose a novel nonlinear disturbance rejection technique for control of robots such as hydraulic robots. Two third-order filters are designed to estimate the system states and torques. The disturbance estimate is calculated by applying the inverse dynamics using the estimated variables. The estimated disturbances are then used by a nonlinear control law to achieve disturbance attenuation control. The proposed technique does not require linearization of the rigid body dynamics, which reduces the computational load. This approach is also less sensitive to noise as will be compared to other existing approaches in literature. 

Finally, the tracking performance and robustness of the proposed approach is validated extensively on real hardware. The robot performs different tasks under different disturbance scenarios (either internal or both internal and external disturbances).




\section{Problem definition}
\label{s:2}
Firstly, this section presents the nominal model of the rigid body dynamics. Then, the model incorporating internal and external disturbances is presented. 
\subsection{Rigid body dynamics}
\label{s:21}
The dynamics of the fixed-base robot can be expressed as
\begin{align}
\label{e:xdot}
\bm{M}(\bm{q}) \ddot{\bm{q}} + \bm{C}(\bm{q},\dot{\bm{q}}) + \bm{G}(\bm{q}) = \bm{\tau}
\end{align}
with $\bm{q}$ is the $n_d\times 1$ vector of joint angular positions with $n_d$ the number of \ac{DOF}, $\bm{M}(\bm{q})$ the ${n_d \times n_d}$ inertia matrix, $\bm{C}(\bm{q},\dot{\bm{q}}) $ the ${n_d}\times 1$ Coriolis and centripetal torques and $\bm{G}(\bm{q})$ the ${n_d}\times 1$ gravitational torques.
For the sake of readability, the dependencies of the system matrices $\bm{M}(\bm{q})$, $\bm{C}(\bm{q},\dot{\bm{q}})$ and $\bm{G}(\bm{q})$ on $\bm{q}$ and $\dot{\bm{q}}$ will be discarded. 
%
%
%
%
\subsection{Rigid body dynamics with internal and external disturbances}
\label{s:22}
For the robot we use in this paper (Fig.~\ref{f:hya}), there are six joints (\ac{SAA}, \ac{SFE}, \ac{HR}, \ac{EFE}, \ac{WR} and \ac{WFE}) \cite{Rehman2015}. Each joint is driven by a hydraulic actuator. The bandwidth of the actuator is dependant on the load it carries \cite{Cunha2013}. For joint \ac{SAA} and \ac{SFE}, the load they carry is higher compared to other joints. Therefore, the actuators have a higher bandwidth. For other joints especially \ac{WR} and \ac{WFE}, the load they carry is significantly smaller than others assuming no external disturbances.

There are static and sliding friction forces in all of the actuators which are internal disturbances. These forces can significantly degrade the torque tracking performance of the actuators. 
For electronic actuators, there are also internal disturbances such as backlash, gearbox friction and static friction \cite{Focchi2013}. In this paper, we model the internal disturbances from the actuators as $\bm{d}_{in}$. 

Apart from internal disturbances, we also consider external disturbances denoted as $ \bm{d}_{ext}$. Considering both internal and external disturbances as well as model uncertainties $\Delta\bm{M}$, $\Delta\bm{C}$ and $\Delta\bm{G}$, the system dynamics Eq.~(\ref{e:xdot}) is rewritten into the following:
\begin{align}
\label{e:xdot uncertain}
\bm{M} \ddot{\bm{q}} + \Delta\bm{M} \ddot{\bm{q}} + \bm{C}+ \Delta\bm{C}  + \bm{G} + \Delta\bm{G}  = \bm{\tau} + \bm{d}_{in} + \bm{d}_{ext}
\end{align}
%
%
Let $\bm{d}$ denote the total disturbances then it follows
\begin{align}
\label{e:xdot d }
\bm{M} \ddot{\bm{q}} +  \bm{C} + \bm{G}  =  \bm{\tau} + \bm{d}
\end{align}
In this paper, the goal is to design a robust controller regardless the presence of $\bm{d}$.

\section{Nonlinear disturbance estimation}
\label{s:3}
The main contribution of this paper is to estimate the disturbance based on the nonlinear dynamic model. From Eq.~\eqref{e:xdot d }, we can obtain the following:
\begin{align}
\label{e:dis}
\bm{d} = \bm{M} \ddot{\bm{q}} +  \bm{C} + \bm{G}  -  \bm{\tau}
\end{align}
We can derive the disturbance estimate as follows:
\begin{align}
\label{e:dis estimate}
\hat{\bm{d}} = \hat{\bm{M}} \hat{\ddot{\bm{q}}} +  \hat{\bm{C}} + \hat{\bm{G}}  -  \hat{\bm{\tau}}
\end{align}
where $\hat{\bullet}$ denotes the estimate.
Let us first calculate $\hat{\ddot{\bm{q}}}$ and $\hat{\bm{\tau}}$.

Most robots have joint encoders which can measure ${\bm{q}}$. We denote the joint angle measurements as ${\bm{q}}_m$. For certain robots, the joint velocity $\dot{\bm{q}}$ is also measured. Angular acceleration sensors are rare, which means that ${\ddot{\bm{q}}}$ is seldomly measured. In our case, only joint encoders, which measure the joint angles, are available. Therefore, we have to estimate both $\dot{\bm{q}}$ and ${\ddot{\bm{q}}}$. Due to the measurement noise, ${\bm{q}}_m$ needs to be filtered in order to get more smooth magnitudes. 

We can estimate ${\bm{q}}$, $\dot{\bm{q}}$ and ${\ddot{\bm{q}}}$ by using the following third-order low-pass filter:
\begin{align}
\label{e:x1_dot}
\dot{\bm{x}}_1 &= {\bm{x}}_2 \\
\dot{\bm{x}}_2 &= {\bm{x}}_3 \\
\label{e:x3_dot}
\dot{\bm{x}}_3 &= ({\bm{q}}_m - \bm{x}_1) *\omega_2^2 \omega_1 \\ \nonumber
 &- ( \omega_2^2 + 2\zeta \omega_2\omega_1 ) \bm{x}_2 - ( 2 \zeta \omega_2 + \omega_1 ) \bm{x}_3
\end{align}
where $
\bm{x}_1 = \hat{\bm{q}}$, $\bm{x}_2 = \hat{ \dot{\bm{q}} }$, $\bm{x}_3 = \hat{ \ddot{\bm{q}} }$. 
The transfer function of this filter is ${\omega_1 \omega_2^2}/{(s+\omega_1) (s^2+2 \zeta \omega_2 s + \omega_2^2) }$.

It can be seen that ${\bm{q}}$, $\dot{\bm{q}}$ and ${\ddot{\bm{q}}}$ are the states and their estimates are obtained from the outputs of the filter. This is the reason why we are using a third-order filter since $\hat{\bm{q}}$, $\hat{ \dot{\bm{q}} }$ and $\hat{ \ddot{\bm{q}} }$ can be directly obtained and filtered by integrating Eqs.~\eqref{e:x1_dot}-\eqref{e:x3_dot}. $\hat{ \ddot{\bm{q}} }$ is obtained without numerical differentiation. Second-order filters can also be used but then there are fewer tuning parameters.

In the dynamics model of the filter, $\zeta$, $\omega_1$ and $\omega_2$ are the damping ratio and natural circular frequencies. $\omega_1$ and $\omega_2$ can be tuned to reduce the delay and reduce the noise effect in ${\bm{q}}_m$. Increasing them means coping with faster disturbance while decreasing them means reducing the noise effect. In reality, if the sensor provides accurate measurement, they can be increased and vice versa.

It can be noted that the input to the system is ${\bm{q}}_m$ and the output is $\bm{x}_1$, $\bm{x}_2$ and $\bm{x}_3$.
The initial value for $\bm{x}_1$, $\bm{x}_2$ and $\bm{x}_3$ can be set according to the initial condition of the robot. The initial value for $\bm{x}_1$ is the initial joint angles before they move. The initial values for $\bm{x}_2$ and $\bm{x}_3$ before the robot moves are set to zero vectors.
By doing this, ${\bm{q}}$, $\dot{\bm{q}}$ and ${\ddot{\bm{q}}}$ are all estimated. 

Although ${\bm{q}}$, $\dot{\bm{q}}$ and ${\ddot{\bm{q}}}$ can all be estimated, there is a phase shift caused by the third-order low-pass filter. In order to cancel the phase shift, we estimate $\hat{\bm{\tau}}$ using the same filter:
\begin{align}
\dot{\bm{y}}_1 &= {\bm{y}}_2 \\
\dot{\bm{y}}_2 &= {\bm{y}}_3 \\
\label{e:y3_dot}
\dot{\bm{y}}_3 &= (\bm{\tau}_{m} - \bm{y}_1) *\omega_2^2 \omega_1 \\ \nonumber
 &- ( \omega_2^2 + 2\zeta \omega_2\omega_1 ) \bm{y}_2 - ( 2 \zeta \omega_2 + \omega_1 ) \bm{y}_3
\end{align}
%
It should be noted that $\zeta$, $\omega_1$ and $\omega_2$ in Eq.~\eqref{e:y3_dot} are the same as the ones in Eq.~\eqref{e:x3_dot}.

Finally, $\hat{\bm{\tau}}$ is obtained by 
\begin{align}
\hat{\bm{\tau}} = \bm{y}_1
\end{align}
The initial value of $\bm{y}_1$ can be set as the initial torque measurement. The initial values of $\bm{y}_2$ and $\bm{y}_3$ can be set to be zero vectors.

After calculating $\hat{\ddot{\bm{q}}}$, we can calculate the inverse dynamics of the robot. It should be noted that $\bm{M}$, $\bm{C}$ and $\bm{G}$ are calculated based on the joint positions ${\bm{q}}$ and velocities $\dot{\bm{q}}$. To avoid phase shift, $\bm{M}$, $\bm{C}$ and $\bm{G}$ should also be calculated according to the estimates $\hat{\bm{q}}$ and $\hat{\dot{\bm{q}}}$ which results in $\bm{M}(\hat{\bm{q}})$, $\bm{C}(\hat{\bm{q}}, \hat{\dot{\bm{q}}})$ and $\bm{G}(\hat{\bm{q}})$ respectively.

Now all terms in Eq.~\eqref{e:dis estimate} are calculated, $\hat{\bm{d}}$ is obtained. In the following section, $\hat{\bm{d}}$ will be used to attenuate the effect of disturbances.

This section proposes a method to estimate the disturbance by designing two third-order filters with pre-designed parameters and then apply the inverse dynamics, which is {\it the main contribution} of this paper. The inverse dynamics can be also efficiently calculated using the method in \cite{Featherstone2008}. 

The proposed method is a nonlinear method, which does not require the linearization of the system model and is thus computationally efficient. Note that this method differs from other filter-based approach from that it uses the filtered version of the torque instead of using the measured torque. The reason why we are doing this is to cancel the phase shift caused by the filter.

Another advantage of this approach is that it does not use $\dot{{\bm{q}}}_m$ as opposed to most disturbance observer-based approaches such as \cite{Chen2000,Mohammadi2013}. We will see the benefit in later sections.

One shortcoming of this approach is that there is delay introduced by the filter. However, this is common in observer-based designs. $\omega_1$ and $\omega_2$ can be increased to reduce the delay. Another solution is to use adaptive Kalman filters \cite{Lu2016a} which reduce delay. But Kalman filters require matrix inversions or sigma points calculations, which increases computational burden. 
\section{Disturbance Attenuation Control Law design}
\label{s:4}
For the controller design, we will use the inverse dynamics control, also known as computed torque control \cite{Sciavicco2005,Craig2005}.

Define the tracking error vector $\bm{e}$ and its derivative $\dot{\bm{e}}$ as
\begin{align}
\label{e:e}
\bm{e} = \bm{q} - \bm{q}_{des}, \dot{\bm{e}} = \dot{\bm{q} } - \dot{\bm{q} }_{des}
\end{align}
where the subscript ``$des$'' denotes the desired value (also called reference value).
For comparison, the control law without disturbance attenuation is presented as follows:
\begin{align}
\label{e:tau des nocomp}
\bm{\tau}_{des} = \bm{M} \ddot{\bm{q}}_{des} + \bm{C}+ \bm{G}  - \bm{K}_D \dot{\bm{e} } - \bm{K}_P \bm{e} 
\end{align}
where $\bm{K}_P$ and $\bm{K}_D$ are the proportional and derivative gains. 

The control law, which makes use of the estimated disturbance for compensation, is as follows:
\begin{align}
\label{e:tau des}
\bm{\tau}_{des} = \bm{M} \ddot{\bm{q}}_{des} + \bm{C}+ \bm{G}  - \bm{K}_D \dot{\bm{e} } - \bm{K}_P \bm{e}  - \hat{\bm{d}}
\end{align}
It should be noted that the calculation of $\bm{M}$, $\bm{C}$ and $\bm{G}$ is based on the current $\bm{q}$ and $\dot{\bm{q}}$. Only $\hat{\bm{d}}$ is calculated based on $\hat{\bm{q}}$ and $\hat{\dot{\bm{q}}}$.

In the following sections, we will compare these two control laws (Eqs.~\eqref{e:tau des nocomp} and \eqref{e:tau des}), as well as other disturbance observers, to show their robustness with respect to disturbances.

\section{Simulation and comparison}
\label{s:}
The performance of the proposed approach will be compared to other techniques such as disturbance observers \cite{Chen2000,Mohammadi2013,Kim2015,Oh2016}. Without losing generality, we will compare the performance of our approach with \cite{Chen2000}. The accuracy of $\hat{\bm{d}}$ is essential in disturbance rejection techniques, we will focus on the comparison of $\hat{\bm{d}}$. 

\begin{figure*}[t]
  \centering
  \subcaptionbox{Real and estimated $\bm{d}$ using \cite{Chen2000}. The standard deviations of noise in $\bm{q}_m$ and $\dot{\bm{q}}_m$ are 0.001 rad and 0.01 rad/s. }[.32\linewidth][c]{%
    \includegraphics[width=.35\linewidth]{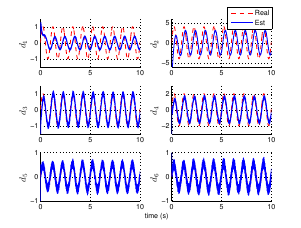}}\quad
  \subcaptionbox{Real and estimated $\bm{d}$ using proposed approach. The standard deviations of noise in $\bm{q}_m$ and $\dot{\bm{q}}_m$ are 0.001 rad and 0.01 rad/s.}[.32\linewidth][c]{%
    \includegraphics[width=.35\linewidth]{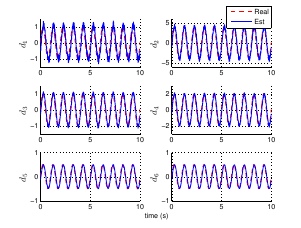}}\quad
  \subcaptionbox{Real and estimated $\bm{d}$ using \cite{Chen2000}. The standard deviations of noise in $\bm{q}_m$ and $\dot{\bm{q}}_m$ are 0.001 rad and 0.1 rad/s.}[.32\linewidth][c]{%
    \includegraphics[width=.35\linewidth]{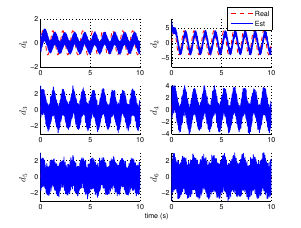}}
    \vspace{-0.0in}
  \caption{Real and estimated $\bm{d}$ using \cite{Chen2000} and proposed approach. The goal is to maintain the robot position in the presence of disturbances which are sine waves with frequency of 1 Hz. Note that the noise-to-signal ratio is increased (compared to sensor used in experiments) to test the performance against noise. }
  \label{f:qd 0.01 0.1 sim}
\end{figure*}

The control objective in this section is to maintain the position of the robotic arm (Fig.~\ref{f:hya}) in the presence of sinusoid disturbances.
In this simulation, we try to cope with ``harsh'' scenarios where the sensor has a high noise-to-signal ratio and the disturbance has a high frequency. Specially, the standard deviations of the noises in $\bm{q}_m$ and $\dot{\bm{q}}_m$ are manually set to 0.001 rad and 0.01 rad/s (note that this is higher than our sensors used in the experiments). The results using \cite{Chen2000} and our approach are shown in Figs.~\ref{f:qd 0.01 0.1 sim}(a) and \ref{f:qd 0.01 0.1 sim}(b).
%
%
%
%
%

It can be readily seen that our approach is robust to noise and can still accurately estimate the disturbances despite the noise-to-signal ratios has been increased. In contrast, \cite{Chen2000} is sensitive to noise and several estimates can not track the real disturbances. If we further increase the standard deviation of $\dot{\bm{q}}_m$ to 0.1 rad/s$^2$, the result using \cite{Chen2000} is shown in Fig.~\ref{f:qd 0.01 0.1 sim}(c). As can be seen, the effect of noise is significantly enlarged. However, our approach is not affected. There are two reasons why \cite{Chen2000} is more sensitive to noise. First of all, they require $\dot{\bm{q}}_m$ which usually has a high noise-to-signal ratio compared to $\dot{\bm{q}}_m$ while our approach do not. The second reason is that observers requires designing a gain which amplifies the noise. It is noted that \cite{Mohammadi2013} proposed to optimize gains to reduce the noise amplification.
%

\section{Experimental Results}
\label{s:6}
In this section, the tracking performance and robustness of the proposed control approach is validated in real hardware. 
The rigid body dynamics is calculated using RobCoGen \cite{Frigerio2016} based on the algorithms of \cite{Featherstone2008}. The parameters are chosen based on the design of the robotic arm shown in Fig.~\ref{f:hya}. The joints are numbered from the shoulder to the wrist.
%

%
For the disturbance estimation, the parameters used in Eq.~\eqref{e:x3_dot} are: $\zeta = 0.8, \omega_1=\omega_2 = 50 \  \text{rad}/\text{s}$.
You can tune $\omega_1$ and $\omega_2$ based on how fast you want to reject the disturbances. If you increase them, the controller can reject faster disturbances. However, it will be illustrated that fast disturbances can also be rejected.

%
In this section, we will not compare with \cite{Chen2000} but rather focus on comparison of control laws Eq.~\eqref{e:tau des nocomp} and ~\eqref{e:tau des} to emphasize the importance of using disturbance rejection techniques for hydraulic robots.
The control law Eq.~\eqref{e:tau des nocomp} is currently implemented for this robotic arm. The controller is implemented in the real time control software SL \cite{schaal2007} and it is run at a rate of 1 KHz. As will be shown, due to friction forces in the actuators, the performance of control law Eq.~\eqref{e:tau des nocomp} is not satisfactory. 

A video of the experiments is provided \footnote{https://youtu.be/TmROclT2ebA}.

To demonstrate the performance, we perform different tasks under different scenarios. 
In Sections~\ref{s:61} and~\ref{s:62}, the tracking performance is validated in the presence of internal disturbances. The task performed in Section~\ref{s:61} is a \textit{goto} task (reach a certain position and maintain the position) while that of Section~\ref{s:62} is a \textit{sine} task (follow a sine shape trajectory for all the joints).
In Sections~\ref{s:63} and~\ref{s:64}, the tracking performance is validated in the presence of both internal and external disturbances. Sections~\ref{s:63} and~\ref{s:64} perform the \textit{goto} task and \textit{sine} task respectively.
%
%
\begin{figure}
\centering
\includegraphics[width = 0.45\textwidth]{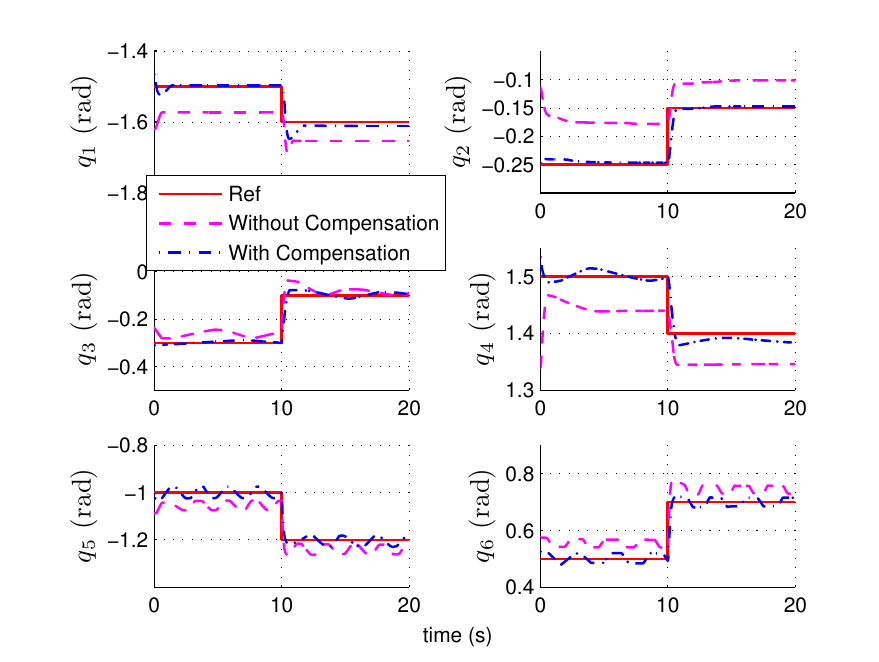}
\vspace{-0.12in}
\caption{Tracking of $q_{des}$ using the controller \textit{without} and \textit{with} disturbance rejection in the \textit{goto} task experiment. The joints are commanded to go to the setpoint at $t=10$ s. The tracking \textit{without} compensation is unsatisfactory while the other one is satisfactory due to the compensation.}
\label{f:q compare real internal}
\end{figure}
\subsection{Experimental validation in the presence of internal disturbance, goto task}
\label{s:61}
In this section, the tracking performance of both control laws will be validated in the presence of internal disturbances (friction forces in the actuators).
The control objective is to reach a setpoint and maintain the position. The joint positions of the setpoint is the current positions plus the incremented positions $[-0.1,0.1,0.2,-0.1,-0.2,0.2]^T$. The start time of the reference is $t=10$ s.

The reference tracking results of the controller without disturbance attenuation are shown in Fig.~\ref{f:q compare real internal}. It is evident that none of the joints reached the desired reference. It is mainly caused by the static friction in the hydraulic actuators. Note that there are some slight oscillations in joint 3, 5 and 6. This is due to the fact that these joints carry less loads which reduces the actuator bandwidth and the force tracking performance of these actuators is worse.

On the contrary, the tracking results of the controller using disturbance attenuation are notably better (shown in Fig.~\ref{f:q compare real internal}). It can be seen that all the friction forces are well compensated. All the joints reach approximately zero-mean tracking performance. Although joint 3, 5 and 6 still have slight oscillations, the biases caused by the friction force are removed, which leads to better tracking.

\begin{figure}
\centering
\includegraphics[width = 0.4\textwidth]{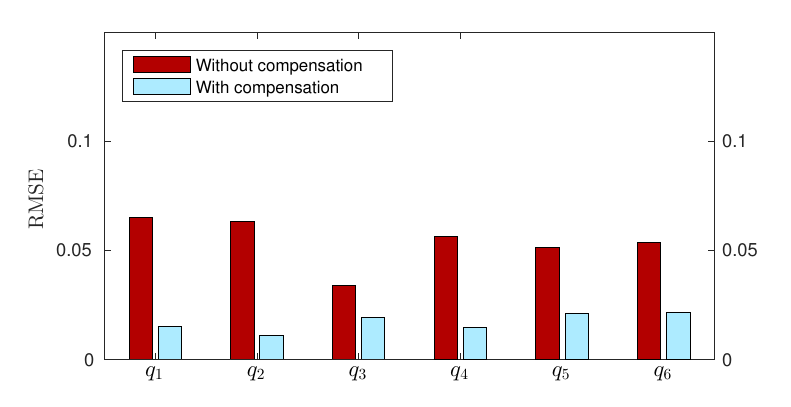}
\vspace{-0.15in}
\caption{Comparison of the \ac{RMSE} of the tracking between two controllers in the \textit{goto} task experiment.}
\label{f:RMSE_q real internal}
\end{figure}

The overall \ac{RMSE} of the tracking using the two controllers are presented in Fig.~\ref{f:RMSE_q real internal}. It can be seen that the \ac{RMSE} using the one with disturbance attenuation are significantly smaller than the other one for all six joints.
%
%
%
%

\begin{figure}
\centering
\includegraphics[width = 0.4\textwidth]{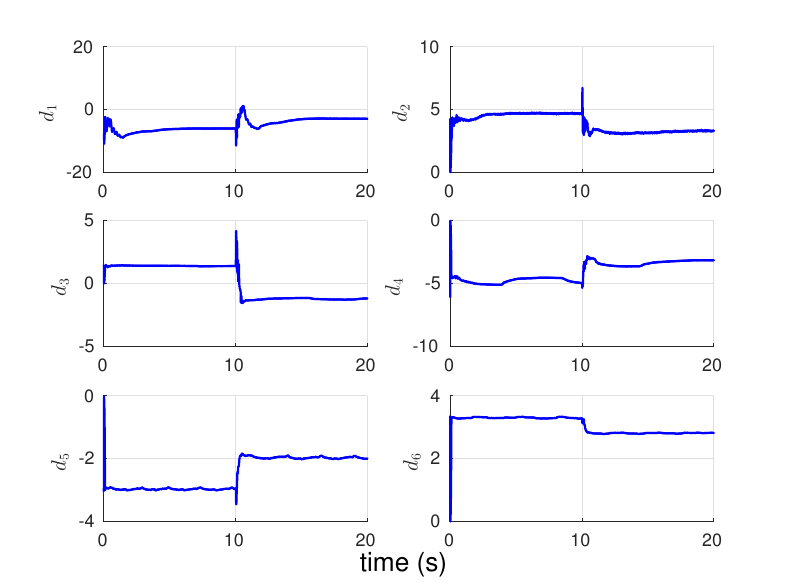}
\vspace{-0.12in}
\caption{Estimate of $\bm{d}$ using the controller with disturbance attenuation in the \textit{goto} task experiment. Note that the disturbances change after the robot reaches the setpoint. }
\label{f:dist real internal}
\end{figure}

Finally, the estimated disturbances $\hat{\bm{d}}$ are given in Fig.~\ref{f:dist real internal}. These internal disturbances are mainly due to the friction in the hydraulic actuators. It is interesting to observe that after the joints reach the setpoint, the internal disturbances changes significantly. This means that the static friction of the actuators changes according to the position. Furthermore, this indicates that on-line disturbance estimation is important since off-line calibration of all joints in all positions is impractical. The performance of the controller using disturbance rejection is significantly better is because these disturbances are estimated on-line and compensated by the control law~\eqref{e:tau des}.
\subsection{Experimental validation in the presence of internal disturbance, sine task}
\label{s:62}
Since our approach is model-based, it can work in various operating points. The objective of this section is to validate the performance in a wider range of motion. To that end, a \textit{sine} task is performed where all the joints follow a sine-shape trajectory. 
%
%
\begin{figure}
\centering
\includegraphics[width = 0.5\textwidth]{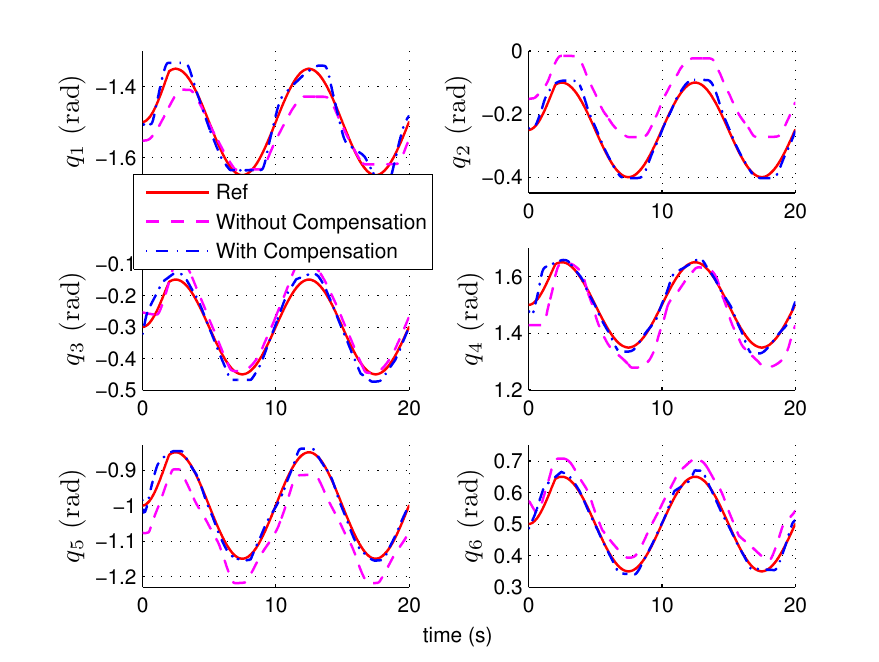}
\vspace{-0.12in}
\caption{Tracking of $q_{des}$ using the controller \textit{without} and \textit{with} disturbance compensation in the \textit{sine} task experiment. For the one \textit{without} compensation, the joint positions significantly deviate from the references while the tracking of the one \textit{with} compensation is significantly better.}
\label{f:q compare real sine internal}
\end{figure}
The reference tracking results of the controller without disturbance attenuation are shown in Fig.~\ref{f:q compare real sine internal}. Differences between the reference (solid lines) and real response (dashed lines) are evident. In contrast, the tracking using the disturbance attenuation (Fig.~\ref{f:q compare real sine internal}) is satisfactory.

\begin{figure}
\centering
\includegraphics[width = 0.4\textwidth]{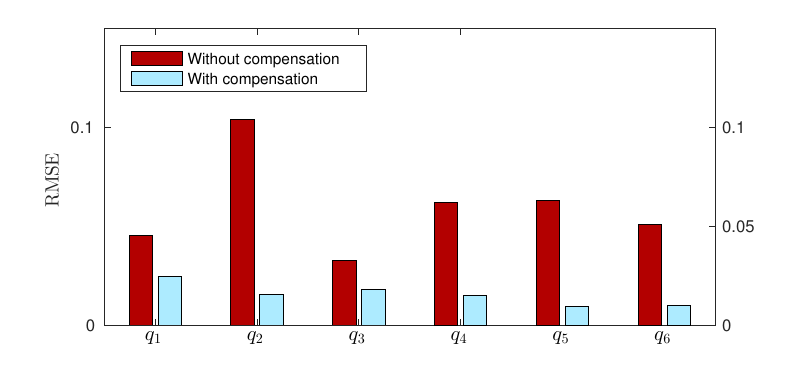}
\vspace{-0.15in}
\caption{Comparison of the \ac{RMSE} of the tracking between two controllers in the \textit{sine} task experiment.}
\label{f:RMSE_q real sine internal}
\end{figure}

The \ac{RMSE} of both controllers are shown in Fig.~\ref{f:RMSE_q real sine internal}. It can be seen that compared to the \textit{goto} task scenario, the errors increased due to the motion of the joints. However, the controller with disturbance rejection excels the other one for all the joints. This again demonstrates the superior performance of the disturbance rejection controller.

\begin{figure}
\centering
\includegraphics[width = 0.4\textwidth]{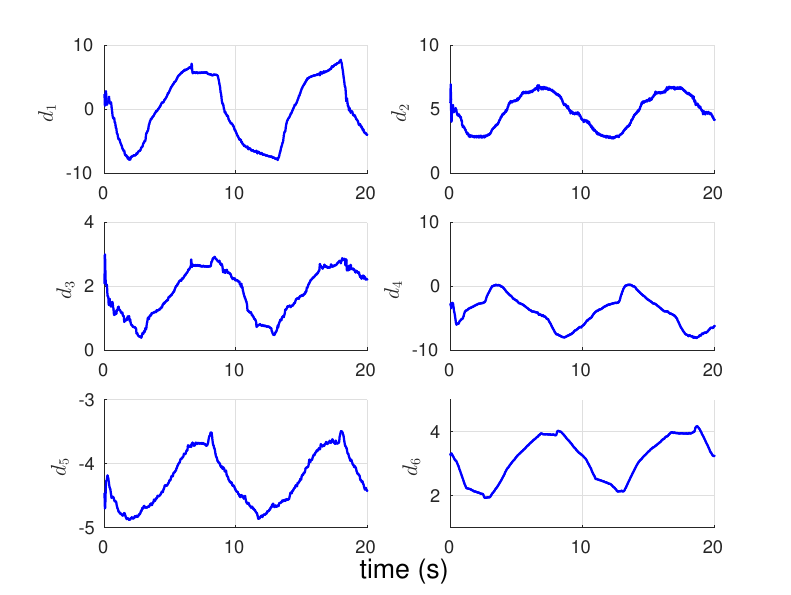}
\vspace{-0.12in}
\caption{Estimate of $\bm{d}$ using the controller with disturbance attenuation in the \textit{sine} task experiment. Note that the direction of the disturbances (friction) are opposite to that of the joint angular velocities.}
\label{f:dist real sine internal}
\end{figure}

The estimated disturbances are shown in Fig.~\ref{f:dist real sine internal}. It is observed that all the disturbances are time-varying and show opposite direction to the joint motion. This is reasonable since the direction of sliding frictions is opposite to the motion of the objects. Beside sliding friction, static friction is also present since the mean of the disturbance is not zero. The bias represents the effects of the static friction.
%
%
%
%
\subsection{Experimental validation in the presence of both internal and external disturbances, goto task}
\label{s:63}
In this and following sections, the tracking performance of both controllers will be validated in the presence of both internal and external disturbances. The scenario of external disturbance is that we suddenly drop a weight to the robot. This is completely unknown to the controller, which means that the weight and when to drop is unknown.
The task performed in this section is \textit{goto} task, same with Section~\ref{s:61}. The only difference is that we will drop a weight on the robot during the execution of the task.
\begin{figure}
\centering
\includegraphics[width = 0.5\textwidth]{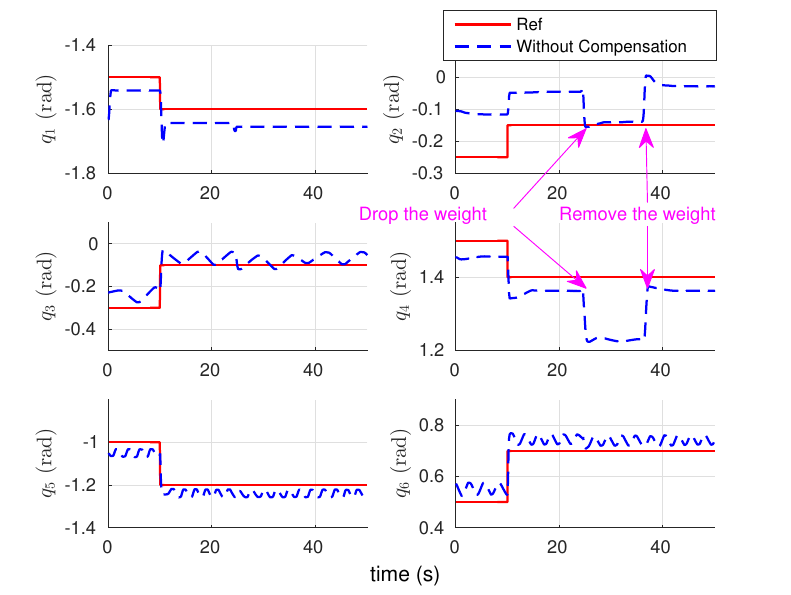}
\vspace{-0.12in}
\caption{Tracking of $q_{des}$ using the controller \textit{without} disturbance rejection in the \textit{goto} task experiment. After the weight is dropped to the end effector, the joint positions change significantly such as the position decreasing of joints 2 and 4. After the weight is removed, the joint positions change again and go back to the positions before the weight is dropped.}
\label{f:q_nocomp real weight}
\end{figure}
\begin{figure}
\centering
\includegraphics[width = 0.5\textwidth]{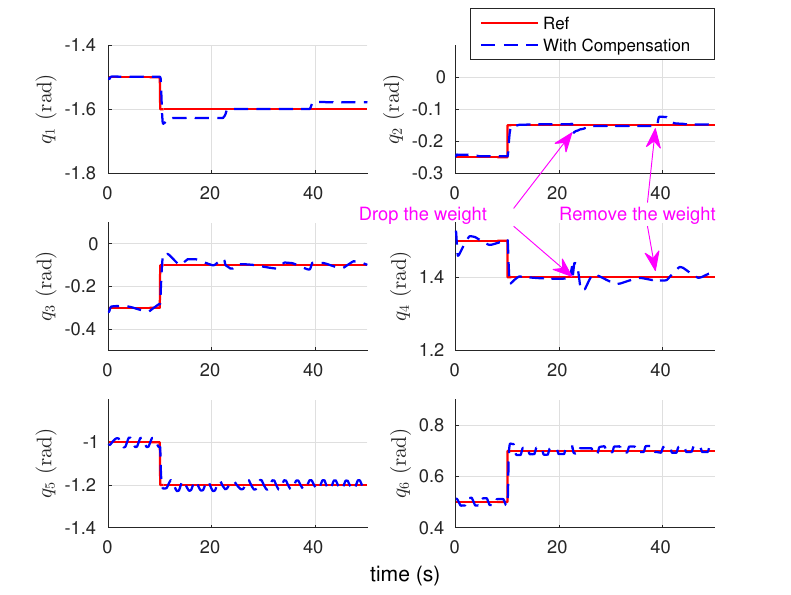}
\vspace{-0.12in}
\caption{Tracking of $q_{des}$ using the controller \textit{with} disturbance rejection in the \textit{goto} task experiment. Whenever the weight is dropped or removed, the joint positions deviate slightly and then quickly recover to the positions when there is no weight.}
\label{f:q real weight}
\end{figure}
%
%

The robot joint position using the controller without disturbance rejection is shown in Fig.~\ref{f:q_nocomp real weight}. At around $t=24.6$ s, the weight is dropped on joint 6 (the end effector). It is clearly seen that after the weight is dropped, the robot position changes significantly. Specifically, joints 2 and 4 drop due to the disturbance. This is because the weight is unknown to the controller and it introduces additional torque in addition to the torque generated by the controller. The additional torque drives the joints going downwards and thus resulting in worse tracking. It is interesting to notice that after the weight is dropped, joint 2 tracks better compared to there is no weight. This is because the directions of the internal disturbance and external disturbance are opposite which cancels the effect of both disturbances. However, after the weight is removed at around $t=36.2$ s, the tracking error increases since there is only internal disturbance.
\begin{figure}
\centering
\begin{minipage}{0.45\columnwidth}
\includegraphics[width = 1\textwidth]{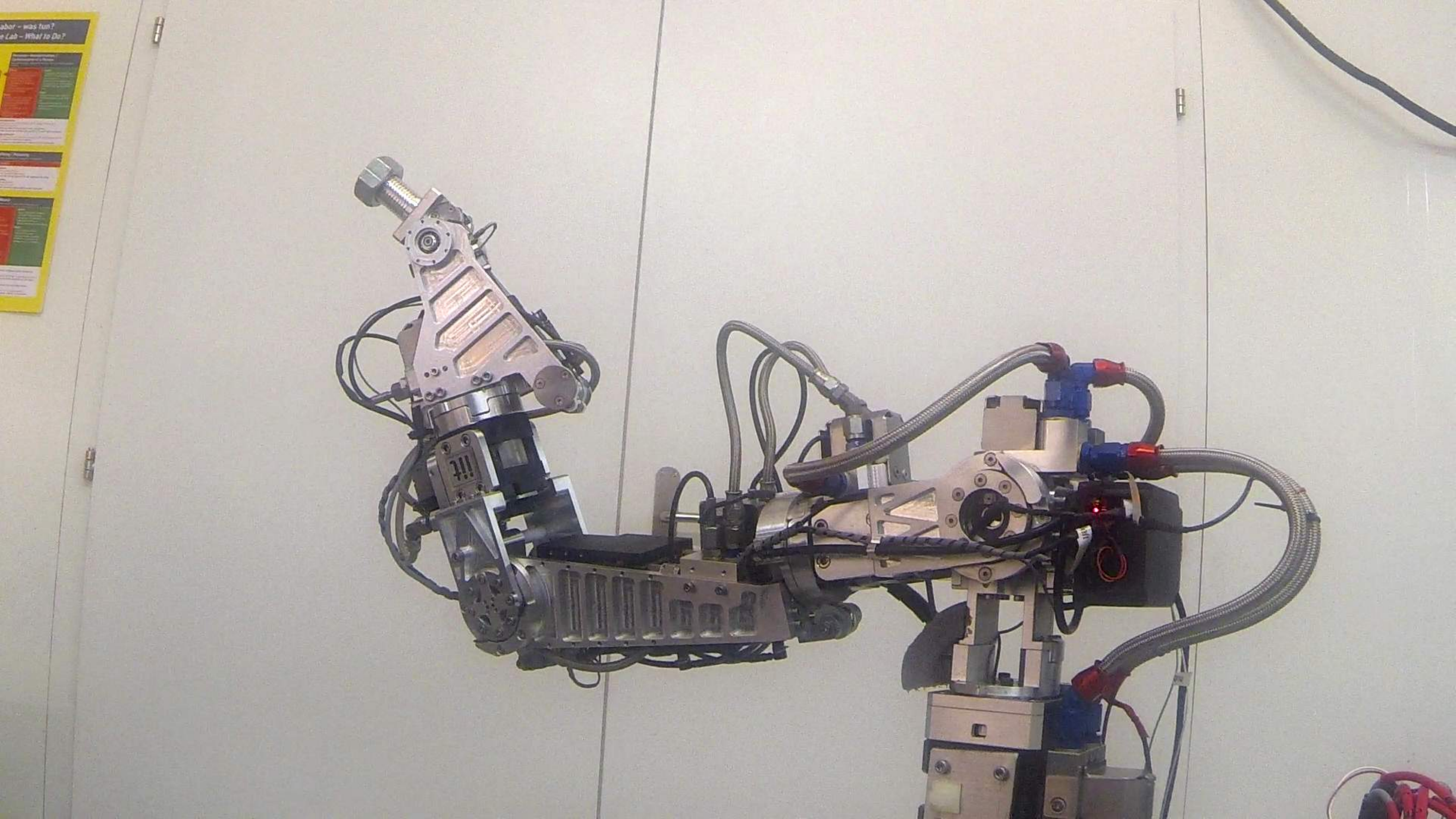}
\vspace{-0.2in}
\caption{Snapshot of the robot \textit{without} disturbance rejection \textit{before} dropping the weight in the \textit{goto} task experiment.}
\label{f:nocomp_before}
\end{minipage}
\begin{minipage}{0.45\columnwidth}
\includegraphics[width = 1\textwidth]{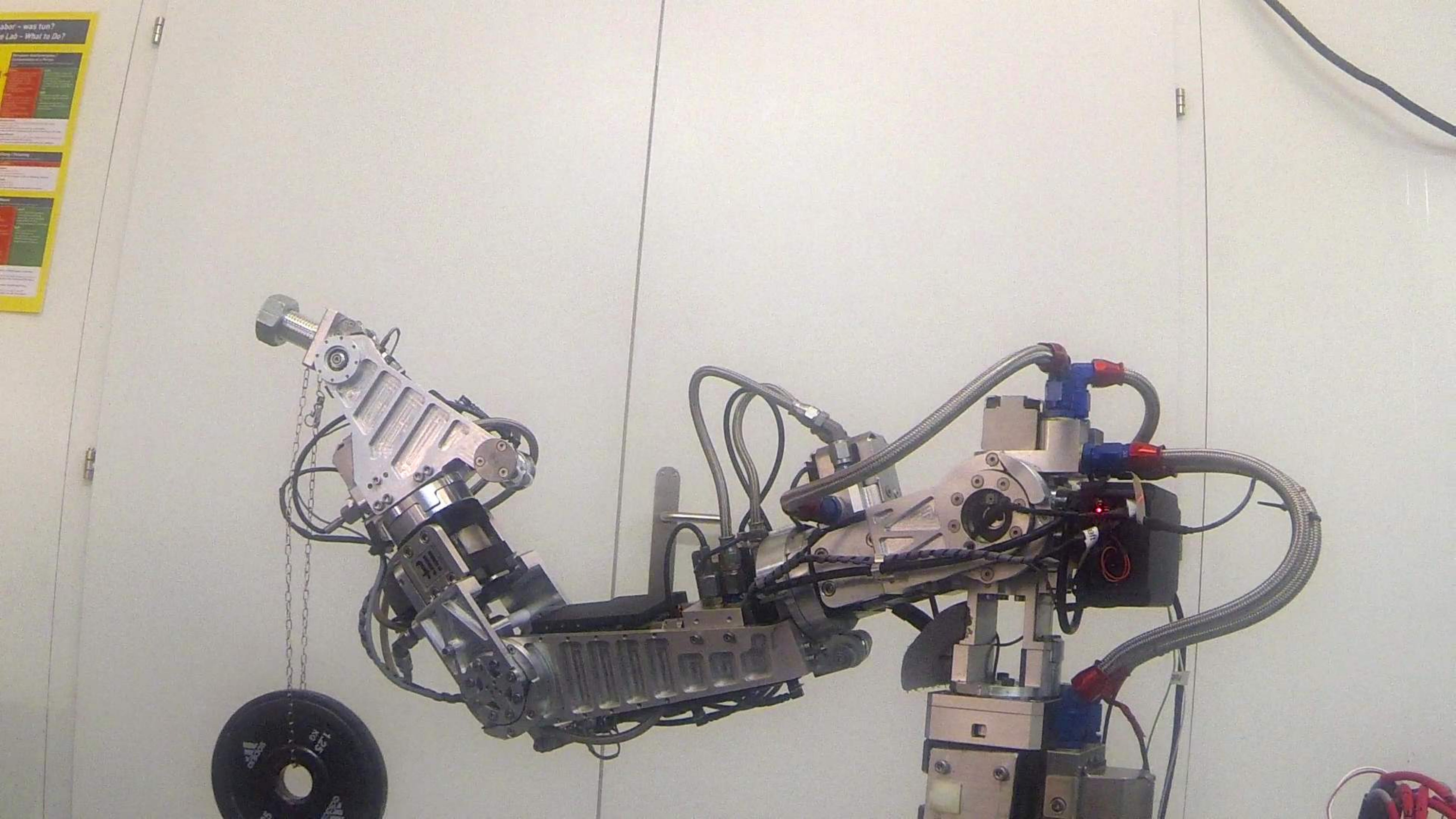}
\vspace{-0.2in}
\caption{Snapshot of the robot \textit{without} disturbance rejection \textit{after} dropping the weight in the \textit{goto} task experiment.}
\label{f:nocomp_after}
\end{minipage}
\end{figure}
The experiment testing external disturbance can be seen in the video. Snapshots of the robot position before and after dropping the weight are shown in Figs.~\ref{f:nocomp_before} and \ref{f:nocomp_after}. From the position of end effector, it can be seen that external disturbance drove the robot downwards.
\begin{figure}
\centering
\begin{minipage}{0.46\columnwidth}
\includegraphics[width = 1\textwidth]{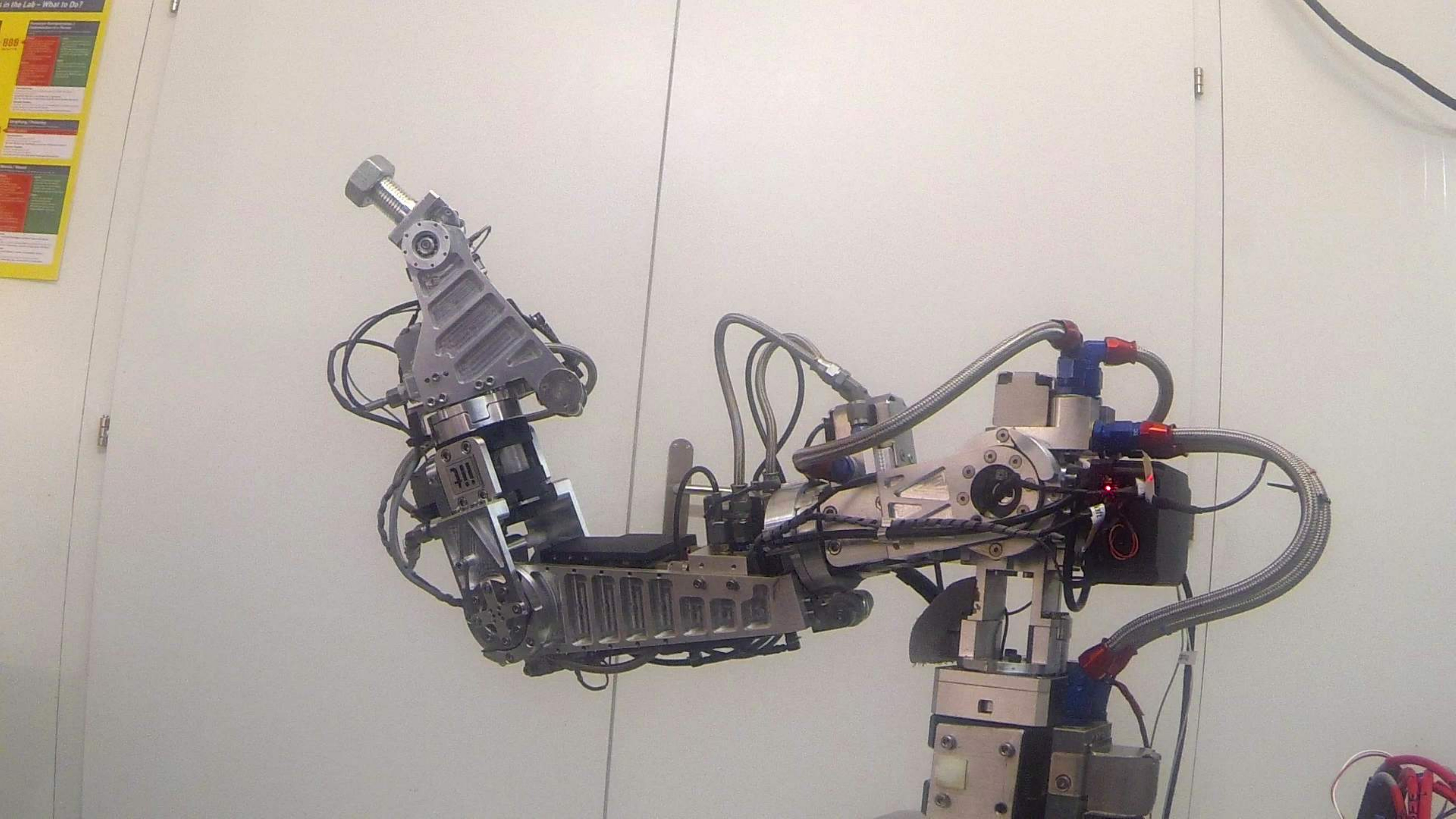}
\vspace{-0.2in}
\caption{Snapshot of the robot \textit{with} the disturbance rejection \textit{before} dropping the weight in the \textit{goto} task experiment.}
\label{f:comp_before}
\end{minipage}
\begin{minipage}{0.45\columnwidth}
\includegraphics[width = 1\textwidth]{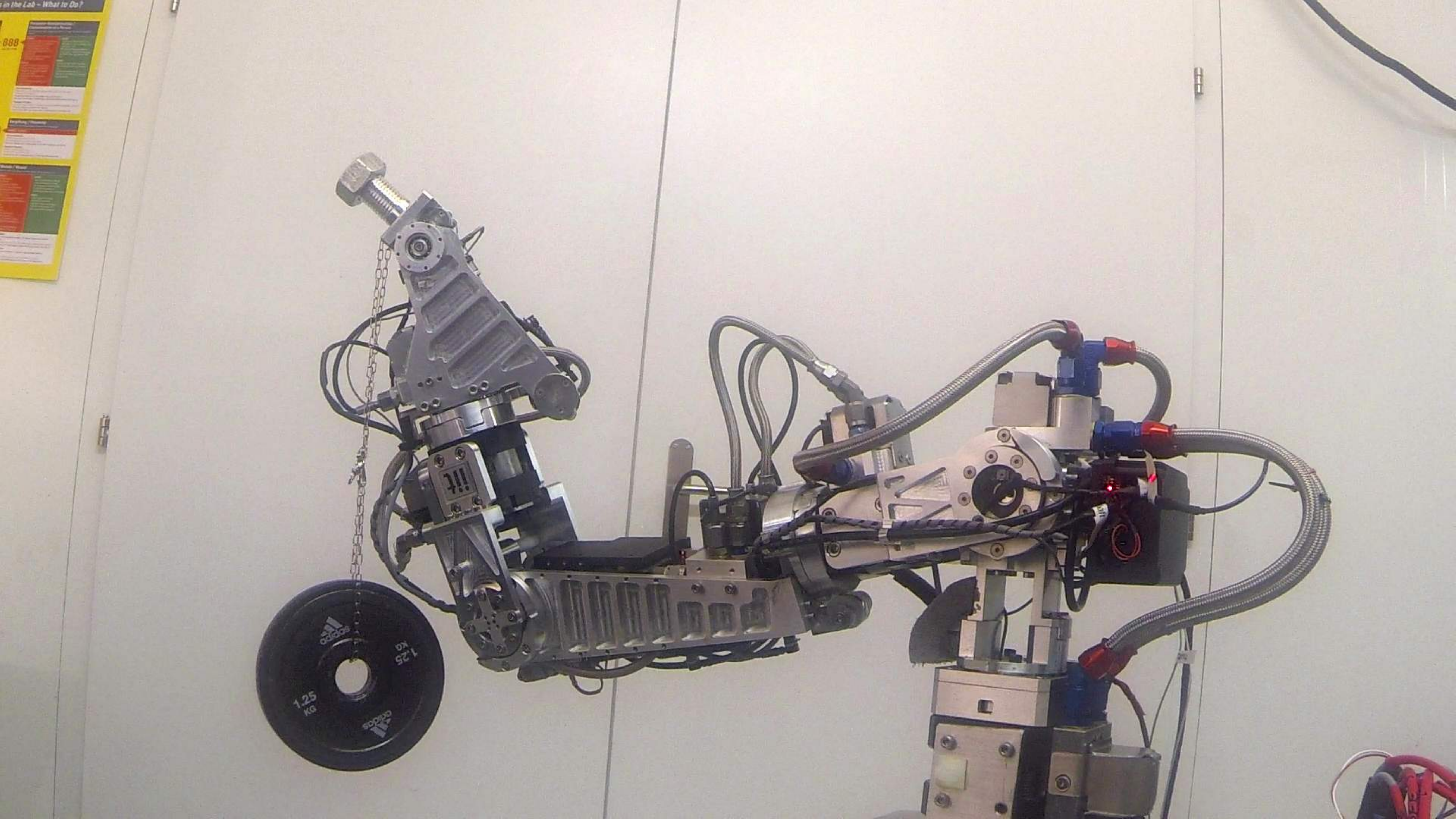}
\vspace{-0.2in}
\caption{Snapshot of the robot \textit{with} the disturbance rejection \textit{after} dropping the weight in the \textit{goto} task experiment.}
\label{f:comp_after}
\end{minipage}
\end{figure}
%
%
%

In contrast, the tracking performance using the controller with disturbance rejection is much less affected by the dropped weight (Fig.~\ref{f:q real weight}). The weight is dropped at around $t=22.5$ s. Once the weight is dropped, the robot joint positions are affected. Then the disturbance rejection fights against the disturbance by compensating for the added weight. It can be seen that although the joint positions change when the weight is dropped (especially joint 2 and 4), they all recovered and continue to maintain satisfactory tracking. After the weight is removed at around $t=39.0$ s, the robot positions changes slightly but immediately recovered to the setpoint. Although the disturbance also changes $\bm{M}$, $\bm{C}$ and $\bm{G}$, they are compensated by the proposed disturbance attenuation approach.
\begin{figure}
\centering
\includegraphics[width = 0.42\textwidth]{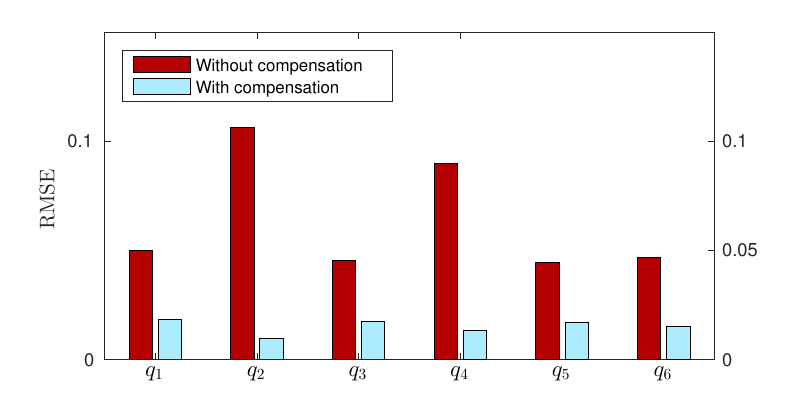}
\vspace{-0.15in}
\caption{Comparison of the \ac{RMSE} of the tracking between two controllers in the \textit{goto} task experiment.}
\label{f:RMSE_q real weight}
\end{figure}

Snapshots of the robot position before and after dropping the weight are shown in Figs.~\ref{f:comp_before} and \ref{f:comp_after}. Observing the end effector positions, it is seen that the position almost remain the same. This demonstrates the robustness (in terms of disturbances) of the proposed approach.
\begin{figure}
\centering
\includegraphics[width = 0.42\textwidth]{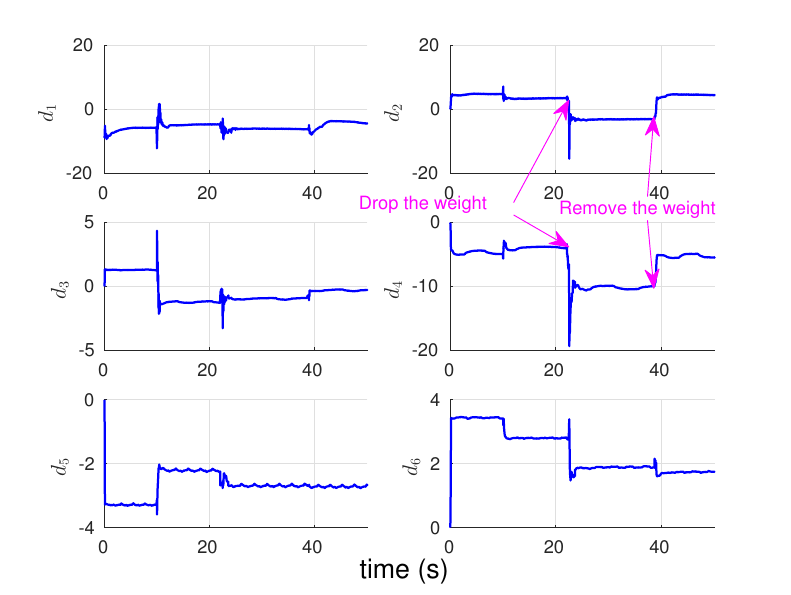}
\vspace{-0.12in}
\caption{Estimate of $\bm{d}$ using disturbance attenuation in the \textit{goto} task experiment. When the weight is dropped at a certain altitude above the end effector at around $t=22.5$ s, it creates momentum besides its own gravity. The total disturbance for joint 4 almost reaches -20 N$\cdot$m.}
\label{f:dist real weight}
\end{figure}
%
%
%

The \ac{RMSE} of the tracking using both controllers are shown in Fig.~\ref{f:RMSE_q real weight}. It is shown that the errors (especially for joint 4) using the controller without disturbance attenuation increase when there are external disturbances compared to Fig.~\ref{f:RMSE_q real internal}. The errors using the controller with disturbance attenuation are smaller than the other one in all joints (especially for joints 2 and 4).

The disturbance estimates are shown in Fig.~\ref{f:dist real weight}. It is seen that when the weight is dropped from a certain altitude to the end effector, it creates a momentum which can be observed from the estimates at around $t=22.5$ s. The internal disturbance in joint 2, which can be observed in the figure before $t=22.5$ s, is around 5 N$\cdot$m. The total disturbance at the moment when the weight is dropped almost reaches -20 N$\cdot$m and maintains at around -5 N$\cdot$m after the weight is dropped. We can derive that the momentum results into -15 N$\cdot$m. The weight creates additional torques in all joints. However, these abrupt disturbances are compensated well by the control law Eq.~\eqref{e:tau des}, the tracking results were shown in Fig.~\ref{f:q real weight}.

\subsection{Experimental validation in the presence of both internal and external disturbances, sine task}
\label{s:64}
In this final section, the tracking performance of both controllers are evaluated during the execution of \textit{sine} task during which the weight will be dropped. 
\begin{figure}
\centering
\includegraphics[width = 0.45\textwidth]{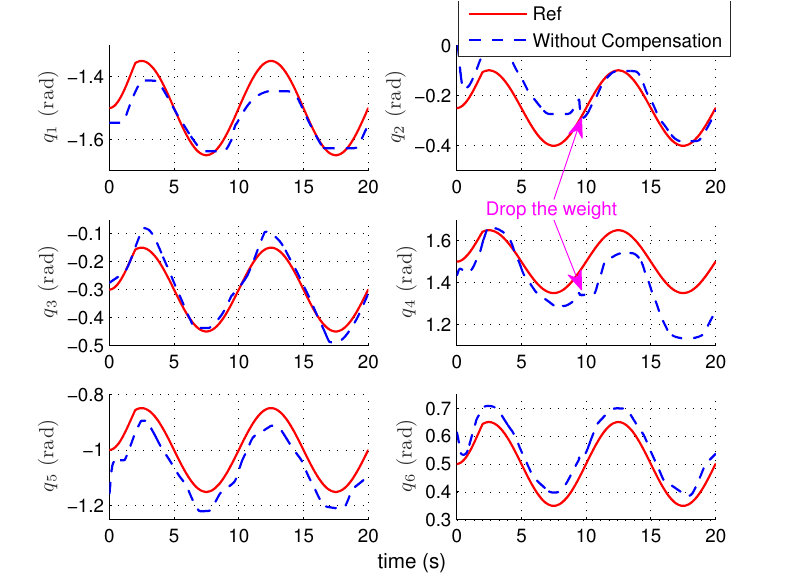}
\vspace{-0.12in}
\caption{Tracking of $q_{des}$ using the controller \textit{without} disturbance attenuation in the \textit{sine} task experiment. After the weight is dropped, the tracking becomes worse especially for joint 4. The reason why tracking of joint 2 becomes better is that the directions of the internal disturbance and the dropped weight is opposite to each other and counteract the effect of both disturbances.}
\label{f:q_nocomp real sine weight}
\end{figure}
\begin{figure}
\centering
\includegraphics[width = 0.45\textwidth]{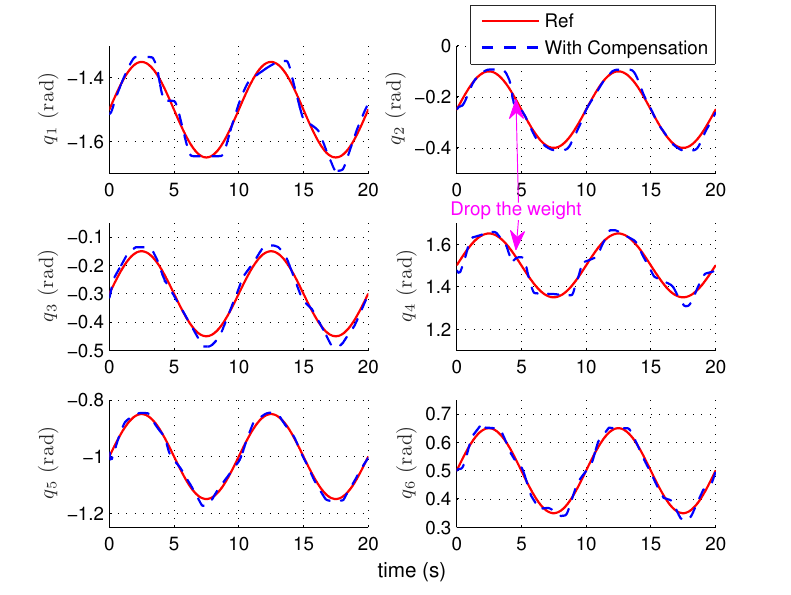}
\vspace{-0.12in}
\caption{Tracking of $q_{des}$ using the controller \textit{with} disturbance attenuation in the \textit{sine} task experiment. The tracking of all joints are satisfactory even after the weight is dropped from a certain altitude.}
\label{f:q real sine weight}
\end{figure}
The joint positions using the controller without and with disturbance rejection are shown in Figs.~\ref{f:q_nocomp real sine weight} and \ref{f:q real sine weight} respectively. Comparing these two figures, it is seen that the one without disturbance rejection could not track the sine reference while the one with disturbance rejection can track the reference better. For Fig.~\ref{f:q_nocomp real sine weight}, it can been that the weight is dropped at around $t=10$ s. However, from Fig.~\ref{f:q real sine weight}, it is difficult to see when the weight is dropped cause the tracking remains the same. 

\begin{figure}
\centering
\includegraphics[width = 0.42\textwidth]{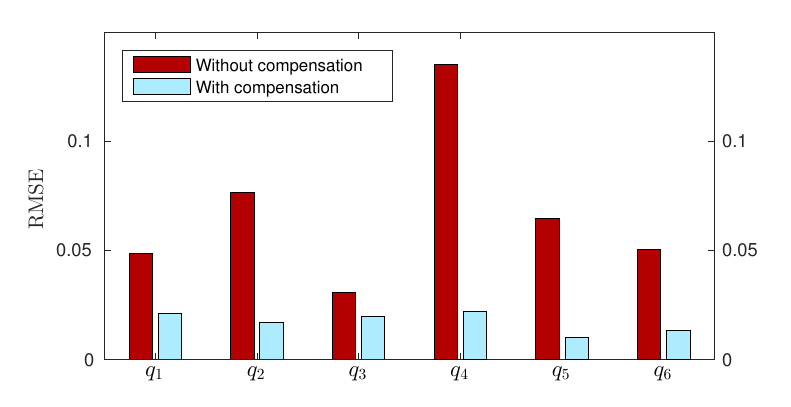}
\vspace{-0.15in}
\caption{Comparison of the \ac{RMSE} of the tracking between two controllers in the \textit{sine} task experiment.}
\label{f:RMSE_q real sine weight}
\end{figure}

The \ac{RMSE} of both controllers are shown in Fig.~\ref{f:RMSE_q real sine weight}. Still, the controller with disturbance rejection outperforms the other one in all joints. This confirms the robustness of the proposed approach since it is always tracking well in the presence or absence of internal or external disturbances.

\begin{figure}
\centering
\includegraphics[width = 0.42\textwidth]{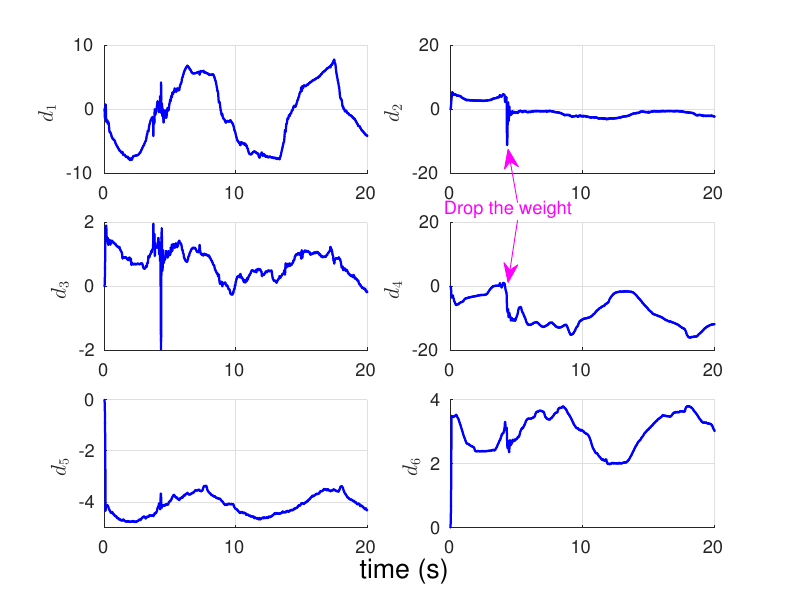}
\vspace{-0.12in}
\caption{Estimate of $\bm{d}$ using the disturbance attenuation in the \textit{sine} task experiment. The disturbances change after the weight is dropped. Since the weight is dropped from a certain altitude, it also creates significant momentum to the joints.}
\label{f:dist real sine weight}
\end{figure}

It is possible to find out when the weight was dropped for the disturbance rejection controller from the disturbance estimate shown in Fig.~\ref{f:dist real sine weight}. It is seen that at around $t=4.2$ s, there is a pulse disturbance, which indicates the dropping of the weight. Almost all joints are affected but joints 2 and 4 are more affected. Even in the presence of this abrupt external disturbance, the controller still tracks well.
\vspace{-0.0in}
\vspace{-0.0in}
%
%
\subsection{Discussions}
Due to the highly nonlinearities and frictions in the hydraulic actuators, it is difficult to track the reference well using inverse dynamic control without disturbance attenuation. The method proposed in this paper works well for the hydraulic robot HyA and can be readily implemented on other hydraulic robotic platforms.

It is worthy to mention that robust control such as chattering-free sliding mode control \cite{Feng2014} did not work on our robotic platform even though we estimate the bound based on the experiments performed in this paper. The possible reason is that the bandwidth of the hydraulic actuator is low such that high gain controllers is difficult to work on this platform. However, this problem deserves more investigation. 

In this paper, we did not look into the dynamics of the hydraulic actuators but rather treat them as a disturbance to the desired torque. The reason is that the dynamics of the hydraulic actuators are highly nonlinear \cite{Boaventura2012} and considering this would lead into more uncertainties. During the operation of the robotic arm which implemented the method in this paper, we did not encounter any problems but it is also worthy to extend our method into the lower level-actuator force control level.

\acresetall
\section{CONCLUSIONS}
\label{s:7}
This paper proposed a nonlinear disturbance attenuation for hydraulic robotic control. The method designs two third-order filters and then applies the inverse dynamics to estimate the disturbances online. The estimated disturbances are used by the controller to achieve disturbance attenuation. The approach is also compared to other existing approaches which demonstrated its superiority. Finally, the proposed approach is validated in four different scenarios either with internal or both internal and external disturbances. The results demonstrate the superior performance and robustness of the proposed approach.
\vspace{-0.10in}


\addtolength{\textheight}{-1cm}   


%

%
%


\bibliography{Robot_control}

\end{document}